\documentclass{article}

\usepackage{microtype}
\usepackage{graphicx}
\usepackage{subfigure}
\usepackage{booktabs} % for professional tables

\usepackage{hyperref}

\usepackage[accepted]{icml2024}

\usepackage{amsmath}
\usepackage{amssymb}
\usepackage{mathtools}
\usepackage{amsthm}
\usepackage{multicol}
\usepackage{multirow}
\usepackage{makecell}
\usepackage{comment}
\usepackage[capitalize,noabbrev]{cleveref}

\theoremstyle{plain}

\theoremstyle{definition}

\theoremstyle{remark}

\usepackage[textsize=tiny]{todonotes}

\begin{document}

\twocolumn[
\icmltitle{ViTAR: Vision Transformer with Any Resolution}



\icmlsetsymbol{equal}{*}

\begin{icmlauthorlist}
\icmlauthor{Qihang Fan}{yyy,comp}
\icmlauthor{Quanzeng You}{sch}
\icmlauthor{Xiaotian Han}{sch}
\icmlauthor{Yongfei Liu}{sch}
\icmlauthor{Yunzhe Tao}{sch}
\icmlauthor{Huaibo Huang}{yyy}
\icmlauthor{Ran He}{yyy,comp}
\icmlauthor{Hongxia Yang}{sch}
\end{icmlauthorlist}

\icmlaffiliation{yyy}{ MAIS \& CRIPAC, Institute of Automa- tion, Chinese Academy of Sciences, China.}
\icmlaffiliation{comp}{ School of Artificial Intelligence, University of Chinese Academy of Sciences, Beijing, China.}
\icmlaffiliation{sch}{ByteDance, Inc.}

\icmlcorrespondingauthor{Qihang Fan}{fanqihang.159@gmail.com}

\icmlkeywords{Machine Learning, ICML}

\vskip 0.3in
]



\printAffiliationsAndNotice  % leave blank if no need to mention equal contribution
{}

\begin{abstract}
This paper tackles a significant challenge faced by Vision Transformers (ViTs): their constrained scalability across different image resolutions.
Typically, ViTs experience a performance decline when processing resolutions different from those seen during training.
Our work introduces two key innovations to address this issue.
Firstly, we propose a novel module for dynamic resolution adjustment, designed with a single Transformer block, specifically to achieve highly efficient incremental token integration. Secondly, we introduce fuzzy positional encoding in the Vision Transformer to provide consistent positional awareness across multiple resolutions, thereby preventing overfitting to any single training resolution.
Our resulting model, ViTAR (Vision Transformer with Any Resolution), demonstrates impressive adaptability, achieving 83.3\% top-1 accuracy at a 1120x1120 resolution and 80.4\% accuracy at a 4032x4032 resolution, all while reducing computational costs.
ViTAR also shows strong performance in downstream tasks such as instance and semantic segmentation and can easily combined with self-supervised learning techniques like Masked AutoEncoder.
Our work provides a cost-effective solution for enhancing the resolution scalability of ViTs, paving the way for more versatile and efficient high-resolution image processing.
\end{abstract}

%This efficient approach allows the ViT to accommodate a wide array of image resolutions while significantly reducing the computational demands associated with high-resolution inputs.
% The Vision Transformer (ViT) is known for its substantial computational demands and limited ability to extrapolate length, which hinders models trained on low-resolution data from being directly transferred to high-resolution scenarios. 
% In this paper, we introduce a plug-and-play cyclical resolution adaptation module that uses just one Transformer block to incrementally integrate tokens. This allows the ViT to accommodate a wide array of image resolutions while requiring considerably less computational resources in high-resolution environments than the original ViT. 
% Moreover, we propose a blurred positional encoding, enabling the Vision Transformer to be aware of position information at any resolution during training, rather than overfitting to a specific resolution. 
% Building upon these two innovations, we designed ViTAR, a model that exhibits commendable performance across a large range of image resolutions (224 to 4000+). 
% Specifically, our ViTAR-B achieves the \textbf{83.4\%} top1 accuracy with the resolution of 896. And the model still has the \textbf{80.4\%} acc when the test resolution is scaled up to 4032. 
% Besides, our ViTAR also has good performance on downstream tasks such as instance segmentation and semantic segmentation. Furthermore, self-supervised learning techniques such as Masked AutoEncoder can be integrated into ViTAR seamlessly.

\section{Introduction}
The tremendous success of the Transformer in the field of Natural Language Processing (NLP) has spurred considerable exploration within the Computer Vision (CV) community~\cite{attention, vit}.
Specifically, Vision Transformers (ViTs) segment images into non-overlapping patches, project each patch into tokens, and then apply Multi-Head Self-Attention (MHSA) to capture the dependencies among different tokens.
Thanks to Transformers' impressive modeling capabilities, ViTs achieve decent results in diverse visual tasks, including image classification~\cite{SwinTransformer, NAT}, object detection~\cite{DETR, pvt, pvtv2}, vision-language modeling~\cite{align, clip}, and even video recognition~\cite{svformer}. 

\begin{figure}
    \centering
    \includegraphics[width=0.99\linewidth]{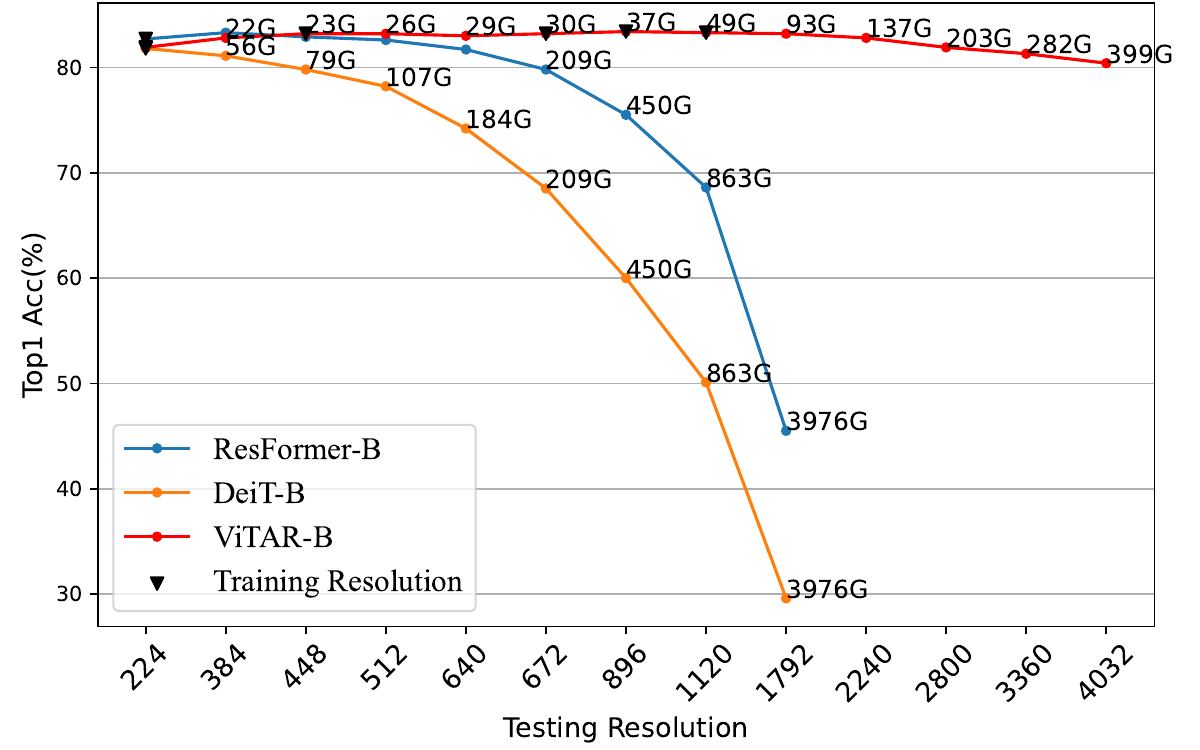}
    \caption{Comparison with other models: When the input resolution is larger than 1792, both DeiT-B and ResFormer-B encounter Out-of-Memory (OOM) errors. Annotations represent the models' computational load in terms of FLOPS. The results demonstrate that ViTAR possesses low computational overhead and exceptionally strong resolution generalization capability.}
    \label{fig:ViTAR_intro}
\end{figure}

Despite achieving success in various domains, ViTs fall short in real-world scenarios that require handling variable input resolutions.
Few studies have explored how to adapting ViTs to different resolutions~\cite{deit, CPVT}. 
Indeed, no training can cover all resolutions, a simple and widely used approach is to directly interpolate the positional encodings before feeding them into the ViT. However, this approach results in significant performance degradation in tasks such as image classification~\cite{deit, resformer, randpos}. To tackle this problem, ResFormer~\cite{resformer} incorporates multiple resolution images during training. Additionally, improvements are made to the positional encodings used by ViT, transforming them into more flexible, convolution-based positional encodings~\cite{CPVT, cswin}. 

However, ResFormer still faces challenges. 
First, it can only maintain a high performance within a relatively narrow range of resolution variations, which is shown in Fig.~\ref{fig:ViTAR_intro}. As the resolution increases, surpassing 892 or even higher, a noticeable decrease in the performance of the model becomes evident. 
Additionally, due to the use of convolution-based positional encoding, it becomes challenging to integrate ResFormer into widely adopted self-supervised frameworks like the Masked AutoEncoder (MAE)~\cite{MAE}.

In this study, we propose the \textbf{Vi}sion \textbf{T}ransformer with \textbf{A}ny \textbf{R}esolution (ViTAR), a ViT which processes high-resolution images with low computational burden and exhibits strong resolution generalization capability. In ViTAR, we introduce the Adaptive Token Merger (ATM) module, which iteratively processes tokens that have undergone the patch embedding. 
ATM scatters all tokens onto grids. 
The process involves initially treating the tokens within a grid as a single unit. 
It then progressively merges tokens within each unit, ultimately mapping all tokens onto a grid of fixed shape. This procedure yields a collection of what are referred to as ``grid tokens''.
Subsequently, this set of grid tokens undergoes feature extraction by a sequence of multiple Multi-Head Self-Attention modules. 
ATM module not only enhances the model's exceptional resolution adaptability but also leads to low computational complexity when dealing with high-resolution images. 
As shown in Fig.~\ref{fig:ViTAR_intro}, our model generalizes better to unseen resolutions compared with DeiT~\cite{deit} and ResFormer~\cite{resformer}. 
Moreover, as the input resolution increases, the computational cost associated with ViTAR is reduced to merely one-tenth or even less than that incurred by the conventional ViT.

In order to enable the model to generalize to arbitrary resolutions, we have also devised a method called Fuzzy Positional Encoding (FPE). The FPE introduces a certain degree of positional perturbation, transforming precise position perception into a fuzzy perception with random noise. This measure prevents the model from overfitting to position at specific resolutions, thereby enhancing the model's resolution adaptability. At the same time, FPE can be understood as a form of implicit data augmentation, allowing the model to learn more robust positional information and achieve better performance.

Our contributions can be summarized as follows:
\begin{itemize}
    \item We propose a simple and effective multi-resolution adaptation module—Adaptive Token Merger, enabling our model to adapt to the requirements of multi-resolution inference. This module significantly improves the model's resolution generalization capability by adaptively merging input tokens and greatly reduces the computational load of the model under high-resolution inputs. 

    \item We introduce a Fuzzy Positional Encoding that allows the model to perceive robust position information during training, rather than overfitting to a specific resolution. We transform the commonly used precise point perception of positional encoding into a Fuzzy range perception. This significantly enhances the model's adaptability to inputs of different resolutions.

    \item We conduct extensive experiments to validate the efficacy of our approach in multi-resolution inference. Our base model attains top-1 accuracies of 81.9, 83.4, and 80.4 under input resolutions of 224, 896, and 4032, respectively. Its robustness significantly surpasses that of existing ViT models. ViTAR also demonstrates robust performance in downstream tasks such as instance segmentation and semantic segmentation.

\end{itemize}

\section{Related Works}

\begin{figure*}[!ht]
    \centering
    \includegraphics[width=0.92\linewidth]{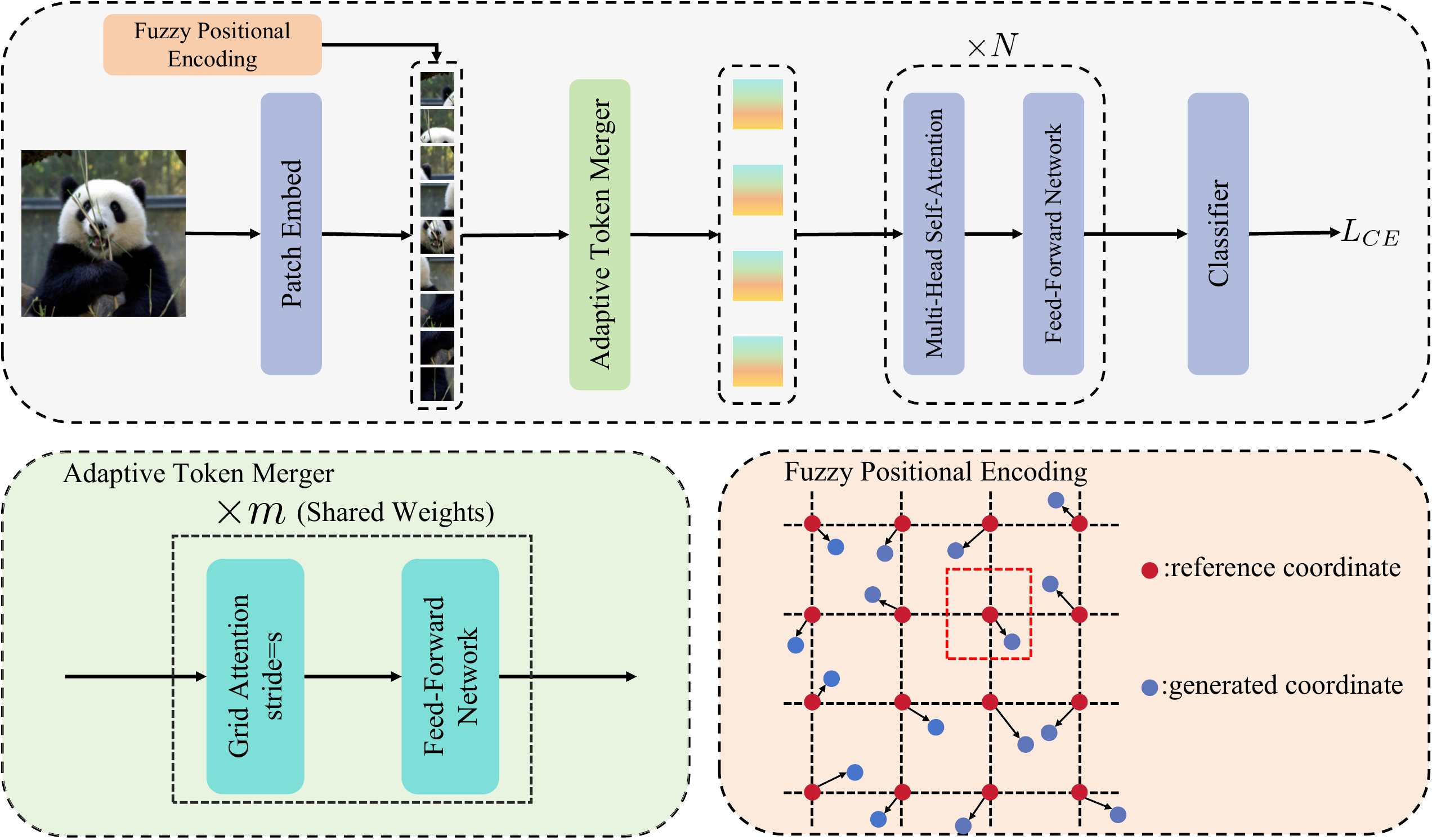}
    \caption{\textbf{Top}: Overall architecture of ViTAR consists of Adaptive Token Merger (ATM), Fuzzy Positional Encoding (FPE), and standard ViT self-attention layers. \textbf{Bottom}: \textit{Left}: ATM with shared weights is applicable $m$ times, depending on the input resolution. \textit{Right}: FPE with the dashed red bounding box to denote the regions from which random points are sampled during training for positional embedding computation.}
    \label{fig:ViTAR}
\end{figure*}

\paragraph{Vision Transformers.}The Vision Transformer (ViT) is a powerful visual architecture that demonstrates impressive performance on image classification, video recognition and vision-language learning~\cite{vit, attention, clip}. Numerous efforts have been made to enhance ViT from the perspectives of data and computation efficiency~\cite{volo, tokenlabel, cmt, FAT, stvit, davit}. In these studies, the majority adapt the model to higher resolutions than those used during training through fine-tuning~\cite{deit, iformer, cswin}. There are few works attempting to directly accommodate unknown resolutions to the model without fine-tuning, which often results in a decline in performance~\cite{SwinTransformer, swinv2}. Fine-tuning on high resolutions often incur additional computational costs. Therefore, designing a visual model that can directly handle multiple resolutions becomes particularly important. However, this direction is still underexplored~\cite{resformer}.

\paragraph{Multi-Resolution Inference.}
Investigating a single visual model that can perform inference across various resolutions remains a largely uncharted field. For the majority of visual models, if the resolution used during inference differs from the ones employed during training and direct inference is conducted without fine-tuning, a performance decline is observed~\cite{SwinTransformer, deit, CPVT}. As pioneering works in this field, NaViT~\cite{NaViT} employs original resolution images as input to ViT, enabling ViT to better handle images with original resolutions without the need for operations such as interpolation that may degrade the image quality. FlexiViT~\cite{flexivit}, on the other hand, uses patches of multiple sizes to train the model, allowing the model to adapt to input resolutions by varying patch sizes. ResFormer~\cite{resformer} employs a method involving multi-resolution training, enabling the model to adapt to input images of various resolutions. It also incorporates several unique positional encodings, enhancing the model's ability to adapt to different resolutions. However, the positional encoding utilized by ResFormer is based on convolutional neural networks~\cite{CPVT, cswin}, and this configuration is challenging to be used in self-supervised learning frameworks like MAE~\cite{MAE}. Additionally, ResFormer itself is based on the original ViT architecture, and when the input resolution increases, it incurs significant computational overhead. To enable the model to adapt to a broader range of resolutions and be applicable to commonly used self-supervised learning frameworks, further model optimization is necessary. 

\paragraph{Positional Encodings.}Positional encoding is crucial for ViT, often providing it with positional awareness and performance improvements. Earlier versions of ViT used sin-cos encoding to convey positional information, and some studies have demonstrated the limited resolution robustness of this positional encoding method~\cite{deit, CPVT, vit}. In contrast, convolution-based positional encoding demonstrates stronger resolution robustness. Models using convolutional positional encoding may even achieve performance improvements when faced with unseen resolutions~\cite{CPVT, resformer, cswin}. Unfortunately, convolutional positional encoding hinders the model to be used in self-supervised learning frameworks such as MAE~\cite{MAE}. This makes it challenging for the model to be employed in the training of large-scale unlabeled datasets~\cite{align}.

\section{Methods}

\subsection{Overall Architecture}

The overall framework of our model is illustrated in Fig.~\ref{fig:ViTAR}, mainly comprising the Adaptive Token Merger (ATM), Fuzzy Positional Encodings (FPE), and the conventional ViT architecture. We do not employ a hierarchical structure; instead, we utilize a straight-through architecture similar to ResFormer~\cite{resformer} and DeiT~\cite{deit}.

\subsection{Adaptive Token Merger (ATM)}

\begin{figure}
    \centering
    \includegraphics[width=0.99\linewidth]{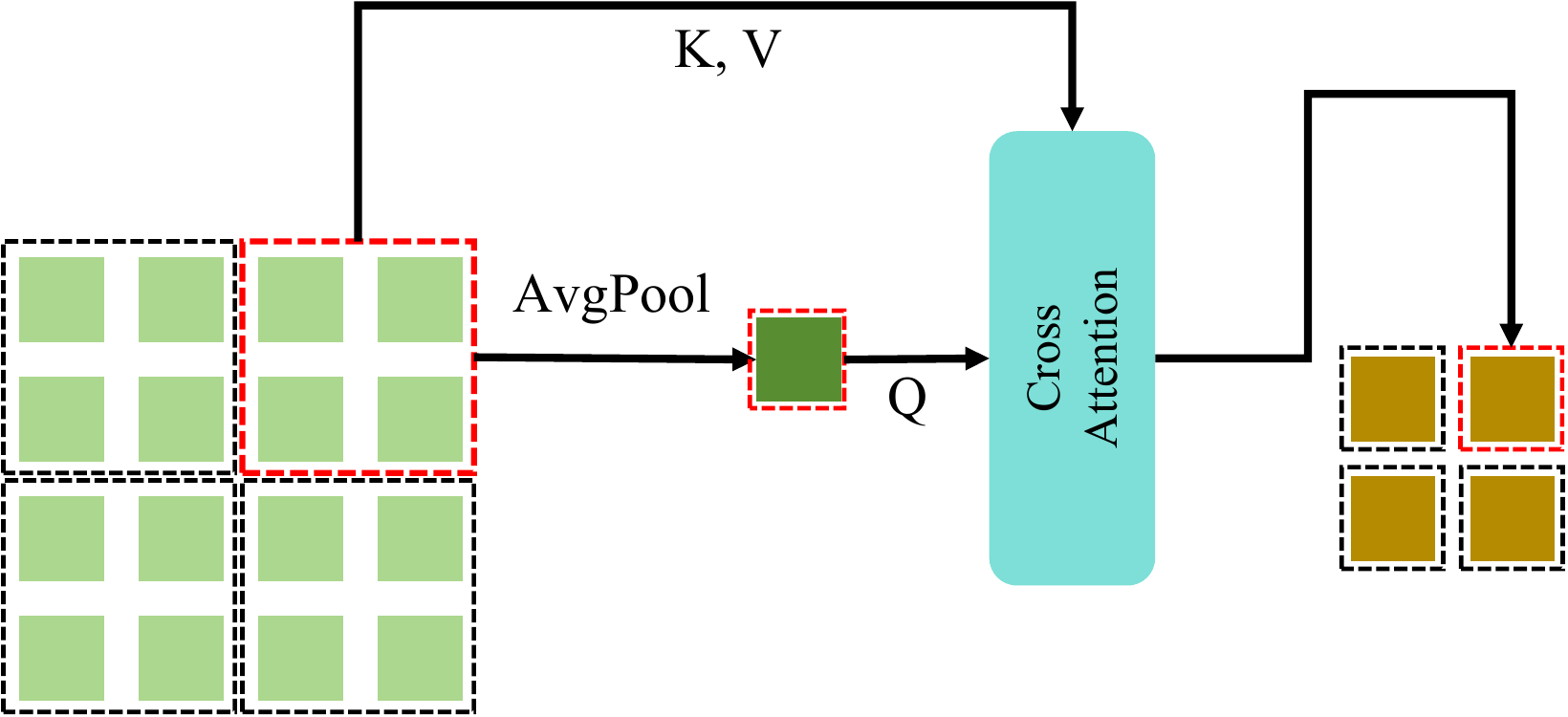}
    \caption{Illustration of GridAttention.}
    \label{fig:gridattention}
\end{figure}

The ATM module receives tokens that have been processed through patch embedding as its input. We preset $G_h\times G_w$ to be the number of tokens we ultimately aim to obtain.
ATM partitions tokens with the shape of $H\times W$ into a grid of size $G_{th}\times G_{tw}$. For ease of representation, we assume $H$ is divisible by $G_{th}$, and $W$ is divisible by $G_{tw}$. Thus, the number of tokens contained in each grid is $\frac{H}{G_{th}}\times \frac{W}{G_{tw}}$. In practical use, $\frac{H}{G_{th}}$ is typically set to 1 or 2, and the same applies to $\frac{W}{G_{tw}}$. Specifically, when $H\ge 2G_h$, we set $G_{th}=\frac{H}{2}$, resulting in $\frac{H}{G_{th}}=2$. When $2G_h>H>G_h$, we pad $H$ to $2G_h$ and set $G_{th}=G_h$, still maintaining $\frac{H}{G_{th}}=2$. The padding method can be found in \textcolor{red}{Appendix}. When $H=G_h$, tokens on the edge of $H$ are no longer fused, resulting in $\frac{H}{G_{th}}=1$.
For a specific grid, suppose its tokens are denoted as $\{x_{ij}\}$, where $0 \le i < \frac{H}{G_{th}}$ and $0 \le j < \frac{W}{G_{tw}}$. We perform average pooling on all $\{x_{ij}\}$ to obtain a mean token. Subsequently, using the mean token as the query, and all $\{x_{ij}\}$ as key and value, we employ cross-attention to merge all tokens within a grid into a single token. We name this process as GridAttention, which can be briefly described using Fig.~\ref{fig:gridattention}. 

Similar to the standard multi-head self-attention, our GridAttention also incorporates residual connections. To align the shapes of tokens, we employ residual connections with average pooling. The complete GridAttention is shown in Eq.~\ref{eq:gridattn}.
\begin{equation}
    \begin{aligned}
        x_{avg}&=\mathrm{AvgPool}(\{x_{ij}\})\\
        \mathrm{GridAttn}(\{x_{ij}\})&=x_{avg}+\mathrm{Attn}(x_{avg}, \{x_{ij}\}, \{x_{ij}\})
    \end{aligned}
    \label{eq:gridattn}
\end{equation}
After passing through GridAttention, the fused token is fed into a standard Feed-Forward Network to complete channel fusion, thereby completing one iteration of merging token. GridAttention and FFN undergo multiple iterations and all iterations share the same weights. During these iterations, we gradually decrease the value of $(G_{th}, G_{tw})$, until $G_{th}=G_h$ and $G_{tw}=G_w$. Similar to the standard ViT, we typically set $G_{h}=G_{w}=14$. This progressive token fusion approach significantly enhances our model's resolution adaptability and reduces the model's computational burden when the input resolution is large.

\subsection{Fuzzy Positional Encoding}

Many studies indicate that commonly used learnable positional encoding and sin-cos positional encoding are highly sensitive to changes in input resolution, and they fail to provide effective resolution adaptability~\cite{CPVT, resformer}. While convolution-based positional encoding exhibits better resolution robustness, its perception of adjacent tokens prevents its application in self-supervised learning frameworks like MAE~\cite{CPVT, MAE}. 

Our FPE differs from the aforementioned methods. While enhancing the model's resolution robustness, it does not introduce a specific spatial structure like convolution does. Therefore, it can be applied to self-supervised learning frameworks. This property enables ViTAR to be applied to large-scale, unlabeled training sets for training, aiming to obtain a more powerful vision foundation model.

In FPE, following ViT, we randomly initialize a set of learnable positional embeddings to serve as the model's positional encoding~\cite{vit}. During the model training, typical positional encodings often provide precise location information to the model~\cite{vit, deit, CPVT}. In contrast, FPE only supplies the model with fuzzy positional information, as illustrated in Fig~\ref{fig:ViTAR}. The positional information provided by FPE undergos shifts within a certain range. Assuming the precise coordinates of the target token is $(i, j)$, the positional coordinates provided by FPE during training are $(i+s_1, j+s_2)$, where $-0.5\le s_1, s_2\le 0.5$. Both $s_1$ and $s_2$ follow a uniform distribution. During the training process, we add randomly generated coordinate offsets to the reference coordinates. Based on the newly generated coordinates, we perform a grid sample on the learnable positional embeddings, resulting in the fuzzy positional encoding. 

During inference, we no longer use the Fuzzy Positional Encoding but opt for the precise positional encoding. When there is a change in the input image resolution, we perform interpolation on the learnable positional embeddings. Due to the use of fuzzy positional encoding during the training phase, for any interpolated positional encoding, the model may have seen and used it somehow. Thus, the model gains robust positional resilience. As a result, during inference, when faced with inputs of unseen resolutions, the model continues to exhibit robust performance.

\subsection{Multi-Resolution Training}
%In contrast to ResFormer, where each batch contains inputs of multiple resolutions and utilizes KL loss for inter-resolution supervision, ViTAR processes each batch with a single resolution input, employing only the basic cross-entropy loss for supervision. 

Similar to ResFormer, when training ViTAR, we also employ a multi-resolution training approach. 
Our model, in contrast to ResFormer~\cite{resformer}, handles high-resolution images with significantly lower computational demands, enabling us to utilize a wider spectrum of resolutions during training.
Unlike ResFormer, which processes batches containing inputs of various resolutions and employs KL loss for inter-resolution supervision, ViTAR processes each batch with inputs of a consistent resolution, relying solely on basic cross-entropy loss for supervision.
Based on the multi-resolution training strategy, our model can be applied to a very broad range of resolutions and achieves favorable results in image classification tasks. Meanwhile, our model achieves similar performance to existing models in high-resolution input tasks (instance segmentation, semantic segmentation) with a much smaller computational cost. Specifically, in instance segmentation and semantic segmentation tasks that require high-resolution inputs, our model achieves results similar to ResFormer and DeiT with 50\% of the FLOPs.

\section{Experiments}

\begin{table*}[ht]
    \centering
    \caption{Comparison on the size ``S'' and size ``B''. Compared to DeiT and ResFormer, ViTAR can handle high-resolution input images with extremely low computational cost and exhibits strong resolution generalization capability. }
    \scalebox{0.9}{
    \begin{tabular}{c|c c|c}
    \toprule[1pt]
         Model & Resolution & FLOPs(G) & Top1-acc(\%)\\
         \midrule[0.5pt]
         DeiT-S & 224 & 5 & 79.8 \\
         ResFormer-S & 224 & 5 & \textbf{82.2}\\
         ViTAR-S & 224 & 5 & 80.3 \\
         %\midrule[0.5pt]
         %DeiT-S & 384 & & 78.1 \\
         %ResFormer-S & 384 & & 
         %ViTAR-S & 384 & 6 & 82.1 \\
         \midrule[0.5pt]
         DeiT-S & 448 & 23 & 75.9 \\
         ResFormer-S & 448 & 23 & 82.5 \\
         ViTAR-S & 448 & 6 & \textbf{82.6} \\
         \midrule[0.5pt]
         DeiT-S & 512 & 32 & 72.6 \\
         ResFormer-S & 512 & 32 & 82.0\\
         ViTAR-S & 512 & 7 & \textbf{82.7} \\
         \midrule[0.5pt]
         DeiT-S & 640 & 58 & 63.9 \\
         ResFormer-S & 640 & 58 & 80.7\\
         ViTAR-S & 640 & 7 & \textbf{82.7} \\
         \midrule[0.5pt]
         %ViTAR-S & 672 & 8 & 82.6 \\
         DeiT-S & 896 & 159 & 41.6 \\
         ResFormer-S & 896 & 159 & 70.8 \\
         ViTAR-S & 896 & 9 & \textbf{83.0} \\
         \midrule[0.5pt]
         DeiT-S & 1120 & 327 & 16.4 \\
         ResFormer-S & 1120 & 327 & 62.7 \\
         ViTAR-S & 1120 & 13 & \textbf{83.1} \\
         \midrule[0.5pt]
         DeiT-S & 1792 & 1722 & 3.5 \\
         ResFormer-S & 1792 & 1722 & 30.8 \\
         ViTAR-S & 1792 & 24 & \textbf{82.8} \\
         \midrule[0.5pt]
         DeiT-S & 2240 & 3965 & 1.5 \\
         ResFormer-S & 2240 & 3965 & 16.1\\
         ViTAR-S & 2240 & 35 & \textbf{82.3} \\
         \midrule[0.5pt]
         DeiT-S & 2800 & OOM & -- \\
         ResFormer-S & 2800 & OOM & -- \\
         ViTAR-S & 2800 & 52 & \textbf{81.3} \\
         \midrule[0.5pt]
         DeiT-S & 3360 & OOM & -- \\
         ResFormer-S & 3360 & OOM & -- \\
         ViTAR-S & 3360 & 72 & \textbf{80.0} \\
         \midrule[0.5pt]
         DeiT-S & 4032 & OOM & -- \\
         ResFormer-S & 4032 & OOM & -- \\
         ViTAR-S & 4032 & 102 & \textbf{78.6} \\
    \bottomrule[1pt]
    \end{tabular}}
    \scalebox{0.9}{
    \begin{tabular}{c|c c|c}
    \toprule[1pt]
         Model & Resolution & FLOPs(G) & Top1-acc(\%)\\
         \midrule[0.5pt]
         DeiT-B & 224 & 18 & 81.8\\
         ResFormer-B & 224 & 18 & \textbf{82.7}\\
         ViTAR-B & 224 & 19 & 81.9 \\
         %\midrule[0.5pt]
         %DeiT-B & 384 & 56 & 81.1\\
         %ResFormer-B & 384 & 56 & 83.3\\
         %ViTAR-B & 384 & 22 & 82.8 \\
         \midrule[0.5pt]
         DeiT-B & 448 & 79 & 79.8\\
         ResFormer-B & 448 & 79 & 82.9 \\
         ViTAR-B & 448 & 23 & \textbf{83.2} \\
         \midrule[0.5pt]
         DeiT-B & 512 & 107 & 78.2\\
         ResFormer-B & 512 & 107 & 82.6 \\
         ViTAR-B & 512 & 26 & 83.2 \\
         \midrule[0.5pt]
         DeiT-B & 640 & 184 & 74.2\\
         ResFormer-B & 640 & 184 & 81.7\\
         ViTAR-B & 640 & 29 & \textbf{83.0} \\
         \midrule[0.5pt]
         %ViTAR-B & 672 & 30 & 83.2 \\
         DeiT-B & 896 & 450 & 64.2 \\
         ResFormer-B & 896 & 450 & 75.5\\
         ViTAR-B & 896 & 37 & \textbf{83.4} \\
         \midrule[0.5pt]
         DeiT-B & 1120 & 863 & 50.1\\
         ResFormer-B & 1120 & 863 & 68.6\\
         ViTAR-B & 1120 & 49 & \textbf{83.3} \\
         \midrule[0.5pt]
         DeiT-B & 1792 & 3976 & 29.6\\
         ResFormer-B & 1792 & 3976 & 45.5\\
         ViTAR-B & 1792 & 93 & \textbf{83.2} \\
         \midrule[0.5pt]
         DeiT-B & 2240 & OOM & -- \\
         ResFormer-B & 2240 & OOM & -- \\
         ViTAR-B & 2240 & 137 & \textbf{82.8} \\
         \midrule[0.5pt]
         DeiT-B & 2800 & OOM & -- \\
         ResFormer-B & 2800 & OOM & -- \\
         ViTAR-B & 2800 & 203 & \textbf{81.9} \\
         \midrule[0.5pt]
         DeiT-B & 3360 & OOM & -- \\
         ResFormer-B & 3360 & OOM & -- \\
         ViTAR-B & 3360 & 282 & \textbf{81.3} \\
         \midrule[0.5pt]
         DeiT-B & 4032 & OOM & -- \\
         ResFormer-B & 4032 & OOM & -- \\
         ViTAR-B & 4032 & 399 & \textbf{80.4} \\
    \bottomrule[1pt]
    \end{tabular}}
    \label{tab:sb}
\end{table*}

We conduct extensive experiments on multiple vision tasks, such as image classification on ImageNet-1K~\cite{imagenet}, instance segmentation on COCO~\cite{coco}, and semantic segmentation on ADE20K~\cite{ade20k}. We also train the model on the self-supervised framework MAE to verify the compatibility between ViTAR and MAE. After them, we make ablation studies to validate the importance of each component in ViTAR. 

\subsection{Image Classification}
\paragraph{Settings.} We train our models on ImageNet-1K~\cite{imagenet} from scratch. We follow the training strategy in DeiT~\cite{deit} but without distillation loss for additional supervision. 
The only supervision for our model is classification loss. We use the AdamW optimizer with a cosine decay learning rate scheduler to train all of our models. 
We set the initial learning rate, weight decay, and batch size to 0.001, 0.05, and 1024, respectively. 
Like DeiT~\cite{deit}, we adopt the strong data augmentation and regularization. 
The settings are RandAugment~\cite{randomaugment} (randm9-mstd0.5-inc1), Mixup~\cite{mixup} (prob=0.8), CutMix~\cite{cutmix} (prob=1.0), Random Erasing~\cite{randera} (prob=0.25). To stabilize the training process, we also use the layer decay~\cite{vit, MAE}.

\paragraph{Results.}The results achieved by the model on ImageNet are shown in Tab.~\ref{tab:sb}. ViTAR demonstrates excellent classification accuracy across a considerable range of resolutions. For models of ``S'' size, when the resolution of input images exceeds 2240, traditional ViT architectures (DeiT and ResFormer) cannot perform inference due to computational resource limitations. In contrast, ViTAR is capable of inference with lower computational costs, maintaining accuracy at a high level (\textbf{2800: 52G, 81.3\%}, \textbf{3360: 72G, 80.0\%}). For models of size ``B'', ViTAR exhibits a similar pattern and adapts well to changes in the input image resolution. It demonstrates the best classification accuracy of \textbf{83.3\%} when the input size is 1120 but only requires 1/20 FLOPs of ViT. Simultaneously, based on experimental results, our model exhibits improved resolution generalization capability as the size increases (at the resolution of 4032, the ViTAR-B achieves a 1.8\% performance improvement compared to the ViTAR-S). 
%As shown in Tab.~\ref{tab:large-res}, we also conduct experiments on a large-sized model (318M). The results validate that our proposed model exhibits excellent scalability.

\begin{comment}
\begin{table}[!ht]
    \centering
    \caption{The adaptability of ViTAR-Large to different resolutions.}
    \begin{tabular}{c|c c|c}
    \toprule[1pt]
         Model & Resolution & FLOPs(G) & Top1-acc(\%)\\
         \midrule[0.5pt]
         ViTAR-L & 224 & 64 & 82.2 \\
         ViTAR-L & 384 & 70 & 82.9 \\
         ViTAR-L & 448 & 71 & 83.4 \\
         ViTAR-L & 512 & 77 & 83.4 \\
         ViTAR-L & 640 & 82 & 83.1 \\
         ViTAR-L & 672 & 83 & 83.6 \\
         ViTAR-L & 896 & 95 & 83.5 \\
         ViTAR-L & 1120 & 117 & 83.5 \\
         ViTAR-L & 1792 & 195 & 83.4 \\
         ViTAR-L & 2240 & 272 & 82.9 \\
         ViTAR-L & 2800 & 389 & 82.5 \\
         ViTAR-L & 3360 & 529 & 82.0 \\
         ViTAR-L & 4032 & 735 & 81.6 \\
    \bottomrule[1pt]
    \end{tabular}
    \label{tab:large-res}
\end{table}
\end{comment}

\subsection{Object Detection}

\paragraph{Settings.} We adopt MMDetection~\cite{mmdetection} to implement Mask-RCNN~\cite{maskrcnn} to validate our model's performance on object detection. The dataset we use is COCO~\cite{coco}. Our experimental settings follow the ResFormer~\cite{resformer} and ViTDet~\cite{vitdet}. We use the commonly used $3\times$ training schedule to train the model. We apply AdamW optimizer with the initial learning rate of $1e-4$ and set the weight decay to $5e-2$. Following previous works~\cite{vitdet, SwinTransformer, pvt}, we use the batch size of 16. 

\begin{table}[!ht]
    \caption{Results and comparisons of different backbones on the COCO2017 with the Mask R-CNN and $3\times$ training schedule. The performances are evaluated by $AP^b$ and $AP^m$.}
    \setlength{\tabcolsep}{0.25mm}
    \centering
    \scalebox{1.0}{
    \begin{tabular}{c|c c c c c c c}
    \toprule[1pt]
         Model & \makecell{Params\\(M)} & $AP^b$ & $AP^b_{50}$ & $AP^b_{75}$ & $AP^m$ & $AP^m_{50}$ & $AP^m_{75}$ \\
         \midrule[0.5pt]
         PVT-Small & 44 & 43.0 & 65.3 & 46.9 & 39.9 & 62.5 & 42.8 \\
         XCiT-S12/16 & 44 & 45.3 & 67.0 & 49.5 & 40.8 & 64.0 & 43.8 \\
         ViT-S & 44 & 44.0 & 66.9 & 47.8 & 39.9 & 63.4 & 42.2\\
         ViTDet-S & 46 & 44.5 & 66.9 & 48.4 & 40.1 & 63.6 & 42.5 \\
         ResFormer-S & 46 & 46.4 & 68.5 & 50.4 & 40.7 & 64.7 & 43.4 \\
         ViTAR-S & 51 & \textbf{46.9} & \textbf{69.0} & \textbf{50.7} & \textbf{41.2} & \textbf{64.9} & \textbf{43.8} \\
         \midrule[0.5pt]
         PVT-Large & 81 & 44.5 & 66.0 & 48.3 & 40.7 & 63.4 & 43.7 \\
         XCiT-M24/16 & 101 & 46.7 & 68.2 & 51.1 &42.0 & 65.6 & 44.9 \\
         ViT-B & 114 & 45.8 & 68.2 & 50.1 & 41.3 & 65.1 & 44.4 \\
         ViTDet-B & 121 & 46.3 & 68.6 & 50.5 & 41.6 & 65.3 & 44.5 \\
         ResFormer-B & 115 & 47.6 & 69.0 & 52.0 & 41.9 & 65.9 & 44.4 \\
         ViTAR-B & 123 & \textbf{48.3} & \textbf{69.6} & \textbf{52.7} & \textbf{42.5} & \textbf{66.7} & \textbf{44.9} \\
         \bottomrule[1pt]
         
    \end{tabular}}
    \label{tab:ob}
\end{table}

\paragraph{Results.} In the tasks of object detection and instance segmentation, we do not utilize the multi-resolution training strategy. Thus, the ATM iterates only once.The $\frac{H}{G_{th}}$ and $\frac{W}{G_{tw}}$ in ATM are fixed to 1. The results, as shown in Tab.~\ref{tab:ob}, indicate that our model achieves excellent performance in both object detection and instance segmentation. Specifically, the $AP^b$ of ViTAR-B exceeds that of ResFormer-B by $0.7$, and ViTAR-B outperforms ResFormer in the remaining metrics. Afterward, to reduce the computational burden of ViTAR, we set $\frac{H}{G_{th}}$ and $\frac{W}{G_{tw}}$ to 2, and compared ViTAR with the standard ViT, as shown in Tab.~\ref{tab:flop}. Our ATM module reduces approximately 50\% of the computational cost while maintaining high precision in dense predictions, demonstrating its effectiveness. 
These outcomes convincingly show that ATM has acquired the capability to adapt to resolutions of various sizes.

%These results strongly demonstrate that ATM has learned knowledge that can adapt to resolutions of any scale.

\begin{table}[!ht]
    \caption{Results and comparison of FLOPs for different backbones. ``*'' indicates that we set $\frac{H}{G_{th}}$ and $\frac{W}{G_{tw}}$ of ATM to 2. FLOPs are measured with the input resolution of $800\times1280$.}
    \setlength{\tabcolsep}{0.3mm}
    \centering
    \scalebox{1.0}{
    \begin{tabular}{c|c c c c c c c}
        \toprule[1pt]
         Model & \makecell{FLOPs\\(G)} & $AP^b$ & $AP^b_{50}$ & $AP^b_{75}$ & $AP^m$ & $AP^m_{50}$ & $AP^m_{75}$ \\
         \midrule[0.5pt]
         ViTDet-B & 812 & 46.3 & 68.6 & 50.5 & 41.6 & 65.3 & 44.5 \\
         ResFormer-B & 823 & 47.6 & 69.0 & 52.0 & 41.9 & 65.9 & 44.4 \\
         ViTAR-B & 836 & \textbf{48.3} & \textbf{69.6} & \textbf{52.7} & \textbf{42.5} & \textbf{66.7} & \textbf{44.9} \\
         ViTAR-B* & \textbf{\textcolor{red}{421}} & \textbf{47.9} & \textbf{69.2} & \textbf{52.2} & \textbf{42.1} & \textbf{66.2} & \textbf{44.5} \\
         \bottomrule[1pt]
    \end{tabular}}
    \label{tab:flop}
\end{table}

\begin{table}[!ht]
    \caption{Results and comparisons of different backbones on ADE20K. All backbones are pretrained on ImageNet-1k.}
    \setlength{\tabcolsep}{1.0mm}
    \centering
    \begin{tabular}{c|c c c c}
    \toprule[1pt]
        Model & \makecell{Params\\(M)} & Lr sched & $\mathrm{mIoU}_{ss}$ & $\mathrm{mIoU}_{ms}$ \\
        \midrule[0.5pt]
        DeiT-S & 52 & 80k & 43.0 & 43.8 \\
        Swin-T & 60 & 160k & 44.5 & 46.1 \\
        XCiT-S12/16 & 52 & 160 & 45.9 & 46.7 \\
        ResFormer-S & 52 & 80k & 46.3 & 47.5 \\
        ViTAR-S & 57 & 80k & \textbf{47.0} & \textbf{48.1} \\
        \midrule[0.5pt]
        DeiT-B & 121 & 160k & 45.4 & 47.2 \\
        XCiT-S24/16 & 109 & 160k & 47.7 & 48.6 \\
        ViT-B+MAE & 177 & 160k & 48.1 & 48.7 \\
        Swin-B & 121 & 160k & 48.1 & 49.7 \\
        ResFormer-B & 120 & 160k & 48.3 & 49.3 \\
        ViTAR-B & 128 & 160k & \textbf{49.0} & \textbf{49.9} \\
        \bottomrule[1pt]
    \end{tabular}
    \label{tab:seg}
\end{table}

\begin{table}[!ht]
    \caption{Results and comparison of FLOPs for different backbones. ``*'' indicates that we set $\frac{H}{G_{th}}$ and $\frac{W}{G_{tw}}$ of ATM to 2. FLOPs are measured with the input resolution of $512\times2048$.}
    \setlength{\tabcolsep}{1.0mm}
    \centering
    \begin{tabular}{c|c c c c}
    \toprule[1pt]
         Model & \makecell{FLOPs\\(G)} & Lr sched & $\mathrm{mIoU}_{ss}$ & $\mathrm{mIoU}_{ms}$ \\
         \midrule[0.5pt]
         %DeiT-S & 1099 & 80k & 43.0 & 43.8 \\
         ResFormer-S & 1099 & 80k & 46.3 & 47.5 \\
         ViTAR-S & 1156 & 80k & \textbf{47.0} & \textbf{48.1} \\
         ViTAR-S* & \textbf{\textcolor{red}{579}} & 80k & \textbf{46.4} & \textbf{47.5} \\
         \midrule[0.5pt]
         %DeiT-B & 1283 & 160k & 45.4 & 47.2 \\
         ViT-B+MAE & 1283 & 160k & 48.1 & 48.7 \\
         ResFormer-B & 1286 & 160k & 48.3 & 49.3 \\
         ViTAR-B & 1296 & 160k & \textbf{49.0} & \textbf{49.9} \\
         ViTAR-B* & \textbf{\textcolor{red}{687}} & 160k & \textbf{48.6} & \textbf{49.4} \\
         \bottomrule[1pt]
    \end{tabular}
    \label{tab:segflops}
\end{table}

\subsection{Semantic Segmentation}
\paragraph{Settings.}Follow the ResFormer, we implement the UperNet~\cite{upernet} with the MMSegmentation~\cite{mmsegmentation} to validate the performance of Our ViTAR. The dataset we use is ADE20K~\cite{ade20k}. To train the UperNet, we follow the default settings in the Swin~\cite{SwinTransformer}. We take AdamW as the optimizer to train the models for 80k/160k iterations.

\begin{table*}[!ht]
    \caption{Results with the MAE framework. The training resolution we used are (224, 448, 672, 896, 1120).}
    \centering
    \begin{tabular}{c c|c c c c c c c c c}
    \toprule[1pt]
         \multirow{2}{*}{Model} & \multirow{2}{*}{Pre-Epochs} & \multicolumn{8}{c}{Testing resolution}\\
         & & 224 & 384 & 448 & 640 & 1120 & 1792 & 2240 & 3360 & 4032 \\
         \midrule[0.5pt]
    ViT-B+MAE & 1600 & \textbf{83.6} & 81.0 & 77.5 & 66.1 & 38.6 & 13.4 & OOM & OOM & OOM \\
    ViTAR-B+MAE & 300 & 82.5 & \textbf{83.2} & \textbf{83.3} & \textbf{83.6} & \textbf{83.5} & \textbf{83.6} & \textbf{83.3} & \textbf{82.6} & \textbf{82.0} \\
    \bottomrule[1pt]
         
    \end{tabular}
    \label{tab:mae}
\end{table*}

\paragraph{Results.} We report the results for segementation in Tab.~\ref{tab:seg} and Tab.~\ref{tab:segflops}. Like what we do in object detection, we first set $\frac{H}{G_{th}}$ and $\frac{W}{G_{tw}}$ of ATM to 1 to compare ViTAR with standard ViT models. As shown in Tab.~\ref{tab:seg}, ViTAR shows better performance than ResFormer and other variants. Specifically, both our Small and Base models outperform ResFormer by 0.6 mIoU. After that, we set the ATM's $\frac{H}{G_{th}}$ and $\frac{W}{G_{tw}}$ to 2, the results are shown in Tab.~\ref{tab:segflops}. Our ATM reduces approximately 40\% of the computational cost while still maintaining accuracy similar to the original model. This strongly demonstrates the effectiveness of the ATM module. The results of instance segmentation and semantic segmentation both indicate that our model can handle high-resolution images with relatively small computational cost, almost without affecting model performance.

\subsection{Compatibility with Self-Supervised Learning}

\paragraph{Settings.}ResFormer employs convolution for positional encoding, making it difficult to be compatible with self-supervised learning frameworks like Mask AutoEncoder (MAE)~\cite{MAE}, which disrupt the spatial structure of images. Since our model does not introduce spatial structures related to convolution, and our proposed Fuzzy Positional Encoding (FPE) does not require additional spatial information, it can be more conveniently integrated into the MAE. Unlike the standard MAE, we still employ a multi-resolution input strategy during training. We pretrain ViTAR-B for 300 epochs and fine-tune for an additional 100 epochs.

\paragraph{Results.}We report the experimental results in Tab.~\ref{tab:mae}. 
ViTAR, pre-trained for just 300 epochs, demonstrates a distinct advantage over a ViT model pre-trained for 1600 epochs.
When the input resolution is increased, ViT+MAE exhibits a significant drop in performance.
On the other hand, ViTAR+MAE demonstrates strong resolution robustness. 
Even with an input resolution exceeding 4000, the model still maintains high performance. 
The findings suggest that our method holds considerable promise for use in self-supervised learning frameworks, as exemplified by MAE~\cite{MAE}.
ViTAR's performance advantage over MAE may stem from two aspects. The first is that ATM enables the model to learn higher-quality tokens, providing the model with a portion of information gain. The second is that FPE, as a form of implicit data augmentation, allows the model to learn more robust positional information. As demonstrated in Droppos~\cite{droppos}, the model's positional information is crucial for its learning process.

\subsection{Ablation Study}

\paragraph{Adaptive Token Merger.}Adaptive Token Merging is a crucial module in our model that enables it to handle multi-resolution inputs. It plays a crucial role in the process of token fusion, effectively reducing the computational burden. Moreover, it enables the model to seamlessly adapt to a range of input resolutions.
We compare two token fusion methods: ATM (Adaptive Token Merger) and AvgPool (adaptive avgpooling). For AvgPool, we partition all tokens into $14\times 14$ grids, and each token within a grid undergoes avgpooling to merge the tokens in that grid. The number of tokens obtained after this fusion is the same as the number of tokens after processing with ATM. The results of the comparison are shown in Tab.~\ref{tab:abatm}, we use the ViTAR-S to conduct the ablation study. The results in Tab.~\ref{tab:abatm} demonstrate that our ATM significantly enhances the performance and resolution adaptability of the model. Especially in high-resolution scenarios, the advantage of ATM becomes increasingly evident. Specifically, at a resolution of 4032, our proposed ATM achieves a 7.6\% increase in accuracy compared with the baseline. At the resolution of 224, ATM also exhibits a performance gain of 0.5\% compared with AvgPool.
\begin{comment}
\begin{table}[ht]
    \caption{Ablation study of ATM. All experiments are conducted based on the ViTAR-S.\qz{if running out of space, make this a 5 columns Table: Method | Res | TOp1-acc || Res | Top1-acc}}
    \centering
    \begin{tabular}{c|c|c}
    \toprule[1pt]
         Method & Resolution & Top1-acc(\%)\\
         \midrule[0.5pt]
         ATM & 224 & 80.3 (\textcolor{red}{+0.5}) \\
         Avg & 224 & 79.8 \\
         \midrule[0.5pt]
         ATM & 448 & 82.6 (\textcolor{red}{+2.5})\\
         Avg & 448 & 80.1 \\
         \midrule[0.5pt]
         ATM & 512 & 82.7 (\textcolor{red}{+3.0})\\
         Avg & 512 & 79.7 \\
         \midrule[0.5pt]
         ATM & 896 & 83.0 (\textcolor{red}{+3.4})\\
         Avg & 896 & 79.6 \\
         \midrule[0.5pt]
         ATM & 1120 & 83.1 (\textcolor{red}{+3.4})\\
         Avg & 1120 & 79.7 \\
         \midrule[0.5pt]
         ATM & 2240 & 82.3 (\textcolor{red}{+6.0})\\
         Avg & 2240 & 76.3 \\
         \midrule[0.5pt]
         ATM & 4032 & 78.6 (\textcolor{red}{+7.6})\\
         Avg & 4032 & 71.0\\
         \bottomrule[1pt]
    \end{tabular}
    \label{tab:abatm}
\end{table}
\end{comment}

\begin{table}[ht]
    \caption{Ablation study of ATM. All experiments are conducted based on ViTAR-S.}
    \centering
    \begin{tabular}{c|c c c c c c}
    \toprule[1pt]
         \multirow{2}{*}{Method} & \multicolumn{5}{c}{Resolution} \\
         & 224 & 448 & 896 & 1120 & 2240 & 4032 \\
         \midrule[0.5pt]
         Avg & 79.8 & 80.1 & 79.6 & 79.7 & 76.3 & 71.0 \\
         ATM & \textbf{80.3} & \textbf{82.6}  & \textbf{83.0} & \textbf{83.1} & \textbf{82.3} & \textbf{78.6} \\
    \bottomrule[1pt]
    \end{tabular}
    \label{tab:abatm}
\end{table}

\paragraph{Fuzzy Positional Encoding.}We compare the impact of different positional encodings on the model's resolution generalization ability. This includes commonly used sin-cos absolute position encoding (APE), conditional position encoding (CPE), global-local positional encoding (GLPE) in ResFormer, Relative Positional Bias (RPB) in Swin~\cite{SwinTransformer}, and our proposed FPE. It is worth noting that only APE and FPE are compatible with the MAE framework. The other two positional encodings, due to the inherent spatial positional structure of convolution, are challenging to integrate into the MAE learning framework. We use the ViTAR-S to conduct the experiments for the models without MAE, and ViTAR-B for the models with MAE. The results of different positional encodings at various test resolutions are shown in Tab~\ref{tab:aFPE}. It can be observed that our proposed FPE exhibits a significantly pronounced advantage in resolution generalization capability. Additionally, under the MAE self-supervised learning framework, FPE also demonstrates superior performance relative to APE, proving the potential applicability of FPE to a broader range of domains. Specifically, at the 4032 input resolution, FPE achieves a top-1 accuracy surpassing GLPE by 4.5\%. In the MAE framework, FPE outperforms APE by 4.6\%.

\begin{table}[!ht]
    \centering
    \caption{Comparison among different methods for positional encodings. Only APE and our FPE can be intergrated into the framework of MAE. }
    \begin{tabular}{c|c c c c c}
    \toprule[1pt]
         \multirow{2}{*}{PE} & \multicolumn{5}{c}{Testing Resolution}\\
         & 224 & 448 & 1120 & 2240 & 4032 \\
         \midrule[0.5pt]
         APE & 80.0 & 81.5 & 81.2 & 76.1 & 70.6 \\
         CPE & 80.4 & 81.8 & 81.6 & 78.2 & 72.0 \\
         RPB & 80.3 & 81.6 & 81.2 & 75.8 & 69.8 \\
         GLPE & \textbf{80.8} & 82.5 & 82.8 & 80.0 & 74.1 \\
         FPE & 80.3 & \textbf{82.6} & \textbf{83.1} & \textbf{82.3} & \textbf{78.6} \\
         \midrule[0.5pt]
         APE+MAE & 82.0 & 82.5 & 82.5 & 80.6 & 77.4 \\
         FPE+MAE & \textbf{82.5} & \textbf{83.3} & \textbf{83.5} & \textbf{83.3} & \textbf{82.0} \\
    \bottomrule[1pt]
    \end{tabular}
    \label{tab:aFPE}
\end{table}

\paragraph{Training Resolutions.}Unlike ResFormer, which only utilizes lower resolutions (128, 160, 224) during training, our model, due to its computational efficiency, can handle the inputs with very high resolutions. 
Additionally, employing a wider range of resolutions enhances the generalization capabilities of our model. In the previous experiments, we use resolutions of (224, 448, 672, 896, 1120) to train all models. In this section, we attempt to reduce the resolutions used during training to examine the model's resolution generalization capability. The experimental results, as shown in Tab.~\ref{tab:abres}, reveal that within the range of resolutions used in the experiments, the model's resolution generalization capability increases with the increment of resolutions used during training. Specifically, when our model is trained using these five resolutions (224, 448, 672, 896, 1120), the model exhibits the strongest resolution generalization capability. It achieves a 4.9\% increase in accuracy at high resolution (4032) compared to training with only (224, 448). This strongly proves the effectiveness of multi-resolution training. 

\begin{table}[ht]
    \centering
    \vspace{-1mm}
    \caption{Ablation of training resolutions. Using more resolutions during training significantly enhances the model's resolution generalization capability. All experiments are conducted based on the ViTAR-S.}
    \setlength{\tabcolsep}{0.8mm}
    \begin{tabular}{c|c c c c c}
    \toprule[1pt]
        \multirow{2}{*}{Training Resolutions} & \multicolumn{5}{c}{Testing Resolutions} \\ 
        & 224 & 448 & 1120 & 2240 & 4032 \\
        \midrule[0.5pt]
        (224, 448, 672, 896, 1120) & 80.3 & 82.6 & 83.1 & 82.3 & 78.6 \\
        (224, 448, 672, 896) & 80.4 & 82.6 & 83.1 & 82.1 & 78.0 \\
        (224, 448, 672) & 80.6 & 82.6 & 83.3 & 81.6 & 76.8 \\
        (224, 448) & 80.7 & 82.8 & 82.2 & 80.2 & 73.7 \\
        \bottomrule[1pt]
        
    \end{tabular}
    \label{tab:abres}
\end{table}

\section{Conclusions}
In this work, we propose a novel architecture: Vision Transformer with Any Resolution (ViTAR). The Adaptive Token Merger in ViTAR enables the model to adaptively handle variable-resolution image inputs, progressively merging tokens to a fixed size, greatly enhancing the model's resolution generalization capability, and reducing computational costs when dealing with high-resolution inputs. Additionally, ViTAR incorporates Fuzzy Positional Encoding, allowing the model to learn robust positional information and handle high-resolution inputs not encountered during training. Our ViTAR is also compatible with existing MAE-based self-supervised learning frameworks, indicating its potential applicability to large-scale unlabeled datasets. In tasks such as instance segmentation and semantic segmentation that require high-resolution inputs, ViTAR significantly reduces computational costs with minimal performance loss. We hope this study can inspire subsequent research on the processing of high-resolution or variable-resolution images.

\bibliography{references}
\bibliographystyle{icml2024}

%%%%%%%%%%%%%%%%%%%%%%%%%%%%%%%%%%%%%%%%%%%%%%%%%%%%%%%%%%%%%%%%%%%%%%%%%%%%%%%
%%%%%%%%%%%%%%%%%%%%%%%%%%%%%%%%%%%%%%%%%%%%%%%%%%%%%%%%%%%%%%%%%%%%%%%%%%%%%%%
% APPENDIX
%%%%%%%%%%%%%%%%%%%%%%%%%%%%%%%%%%%%%%%%%%%%%%%%%%%%%%%%%%%%%%%%%%%%%%%%%%%%%%%
%%%%%%%%%%%%%%%%%%%%%%%%%%%%%%%%%%%%%%%%%%%%%%%%%%%%%%%%%%%%%%%%%%%%%%%%%%%%%%%
\newpage
\appendix
\onecolumn
\section{Grid Padding.}

\begin{figure}[h]
    \centering
    \includegraphics[width=0.7\linewidth]{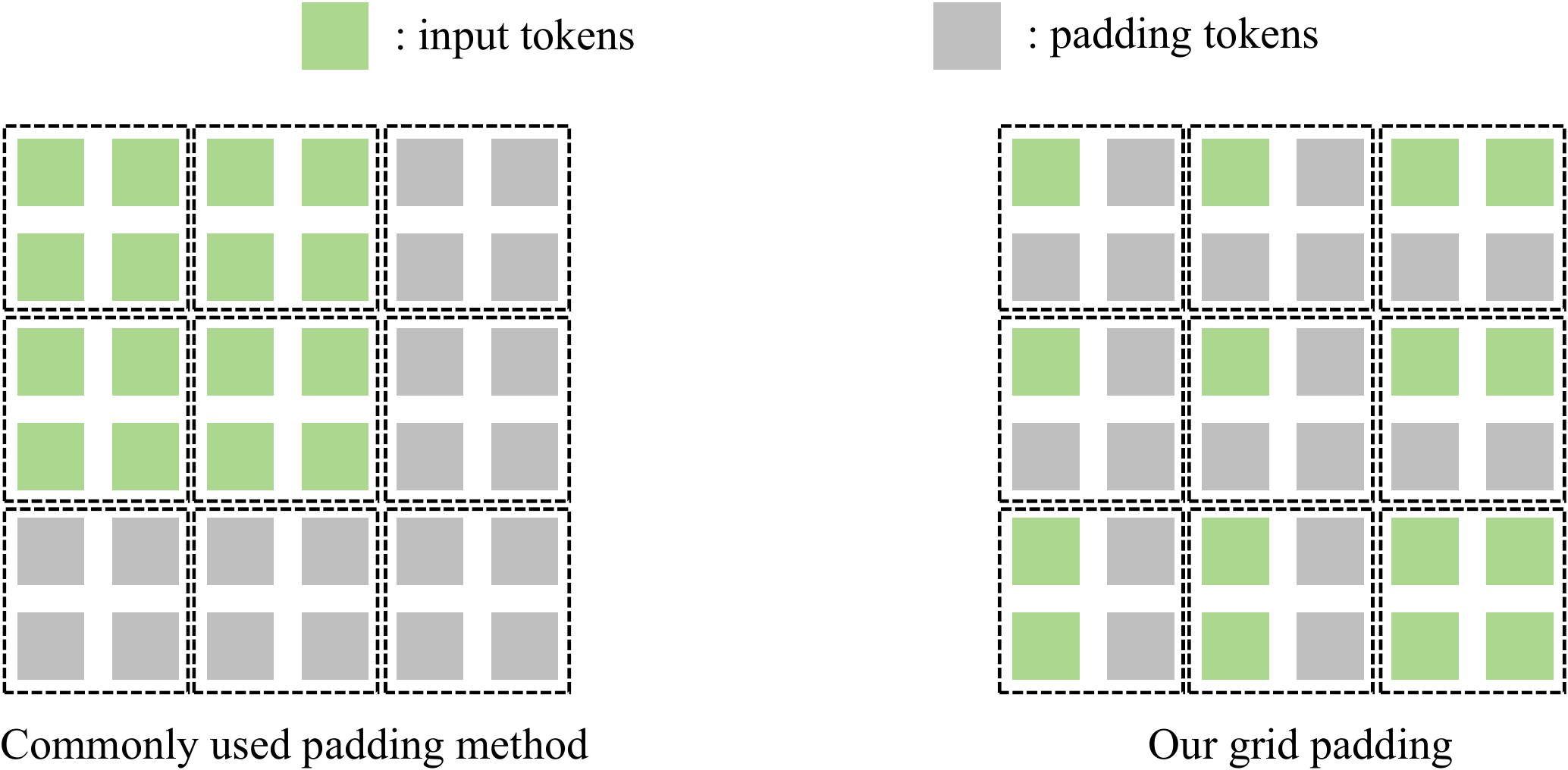}
    \caption{Illustration of grid padding.}
    \label{fig:gridpadding}
\end{figure}

The commonly used padding approach involves surrounding the original tokens with an additional layer of padding tokens. However, in the context of ATM, employing this method might result in the outer grids containing only padding tokens, rendering the grid tokens ineffective. Therefore, as depicted in Fig.~\ref{fig:gridpadding}, we employ the grid padding method. Specifically, we pad with additional tokens on the right and bottom sides of each grid, ensuring a uniform distribution of padding tokens within each grid. This prevents the occurrence of ineffective grids.

%%%%%%%%%%%%%%%%%%%%%%%%%%%%%%%%%%%%%%%%%%%%%%%%%%%%%%%%%%%%%%%%%%%%%%%%%%%%%%%
%%%%%%%%%%%%%%%%%%%%%%%%%%%%%%%%%%%%%%%%%%%%%%%%%%%%%%%%%%%%%%%%%%%%%%%%%%%%%%%

\end{document}